\documentclass[journal]{IEEEtran}
\usepackage[colorlinks,linkcolor=red]{hyperref}

\usepackage{amsmath}
\usepackage{graphicx}
\usepackage{booktabs} 
\usepackage{amssymb}
\usepackage{float}
\usepackage{comment}
\usepackage{mathrsfs}
\usepackage{color}
\usepackage{xcolor}
\usepackage{subcaption}
\usepackage{cite}

\ifCLASSINFOpdf
\else
\fi
\begin{document}
%
\title{DATransNet: Dynamic Attention Transformer Network for Infrared Small Target Detection}
%
%
%

\author{Chen Hu, Yian Huang,~\IEEEmembership{Student Member,~IEEE}, Kexuan Li, Luping Zhang, Chang Long, Yiming Zhu, Tian Pu, and Zhenming Peng,~\IEEEmembership{Member,~IEEE}

\thanks{Manuscript received XXX XXX, XXX; revised XXX XXX, XXX.}
\thanks{This work was supported by Natural Science Foundation of Sichuan Province of China (Grant No. 2025ZNSFSC0522) and partially supported by National Natural Science Foundation of China (Grant No.61775030, Grant No.61571096). \textit{(Corresponding authors: Zhenming Peng; Tian Pu.)}}
\thanks{Chen Hu, Yian Huang, Kexuan Li, Chang Long, Luping Zhang, Tian Pu, and Zhenming Peng are with the Laboratory of Imaging Detection and Intelligent Perception and the School of Information and Communication Engineering, University of Electronic Science and Technology of China, Chengdu 610054, China (e-mail: 202221011506@std.uestc.edu.cn; huangyian@std.uestc.edu.cn; 202322011823@std.uestc.edu.cn; anguing@foxmail.com; lc243265379@gmail.com; putian@uestc.edu.cn; zmpeng@uestc.edu.cn)
}
\thanks{Yiming Zhu is with School of Science, Hangzhou Dianzi University, Hangzhou, China. 
    (e-mail:yiming\_zhu\_grokcv@163.com)
}
\thanks{}

}

\markboth{Journal of \LaTeX\ Class Files,~Vol.~14, No.~8, August~2015}%
{Shell \MakeLowercase{\textit{et al.}}: Bare Demo of IEEEtran.cls for IEEE Journals}
%



\maketitle
\begin{abstract}

Infrared small target detection (ISTD) is widely used in civilian and military applications. 
However, ISTD encounters several challenges, including the tendency for small and dim targets to be obscured by complex backgrounds.
To address this issue, we propose the Dynamic Attention Transformer Network (DATransNet), which aims to extract and preserve detailed information vital for small targets.
DATransNet employs the Dynamic Attention Transformer (DATrans), simulating central difference convolutions (CDC) to extract gradient features.
Furthermore, we propose a global feature extraction module (GFEM) that offers a comprehensive perspective to prevent the network from focusing solely on details while neglecting the global information. 
We compare the network with state-of-the-art (SOTA) approaches and demonstrate that our method performs effectively. 
Our source code is available at \url{https://github.com/greekinRoma/DATransNet}.


\end{abstract}

\begin{IEEEkeywords}
Infrared small target detection (ISTD), convolution neural network (CNN), Dynamic Attention Transformer, global feature extraction.
\end{IEEEkeywords}

%
\IEEEpeerreviewmaketitle

\section{Introduction}

\IEEEPARstart{I}{NFRARED} small target detection (ISTD) is vital for various fields. Currently, ISTD methods can be categorized into model-driven and data-driven methods.


Model-driven methods include three main approaches. 1) Filter-based methods, such as the Tophat \cite{tom1993morphology}. 2) Methods based on the human visual system (HVS), such as the local contrast measure (LCM)\cite{chen2013local} and multi-patch contrast measure (MPCM)\cite{wei2016multiscale}. 3) Low-rank matrix decomposition and reconstruction methods, such as infrared patch image (IPI)\cite{gao2013infrared}, and reweighted infrared patch tensor (RIPT) \cite{dai2017reweighted}.

With the advancement of deep learning, data-driven approaches have achieved substantial progress in ISTD. For example, the Asymmetric Context Modulation (ACM) network \cite{dai2021asymmetric} introduces asymmetric feature fusion, an alternative to conventional skip connections in U-Net. The dense nested attention network (DNANet) \cite{li2022dense} implements a multilayer nested architecture that supports progressive and adaptive interactions between feature layers. Moreover, UIUNet \cite{wu2022uiu} enhances the detection of local target contrasts by integrating multiple U-Net structures and using interactive cross-attention mechanisms for feature fusion.
Additionally, Gated-shaped TransUnet (GSTUnet) \cite{zhu2024towards} merges Vision Transformer with CNNs in the encoder to learn both global and local information. 
 Receptive-field and Direction-induced Attention Network (RDIAN) \cite{sun2023receptive} utilizes different receptive fields and multi-direction-guided attention to enhance the features of targets.
Attentional Local Contrast Network (ALCNet) \cite{dai2021attentional} designs a bottom-up attention modulation to strengthen the small target characteristics.
Attention-guided Pyramid Context Network (AGPCNet) \cite{zhang2023attention} divides the image into several patches and computes both global and local associations.
Yuan et al. \cite{Yuan2024SCTransNetSC} propose the Spatial-channel Cross Transformer Network (SCTransNet), which uses the Spatial-channel Cross Transformer Blocks (SCTBs) to improve the capacity of global information modelling.
Generally, these data-driven methods surpass traditional model-driven approaches.

Although data-driven methods achieve excellent performance, they often struggle with limited ability to capture details and weak global perceptual capabilities. To address these limitations, we propose a novel framework called the Dynamic Attention Transformer Network (DATransNet). 
DATransNet incorporates two key modules. 
We first introduce the Dynamic Attention Transformer (DATrans), which is good at capturing details.  
Then, the Global Feature Extractor Module (GFEM) gives our model a global perspective over the whole image. 


The main contributions of this letter are as follows.

\begin{figure*}
    \centering
    \includegraphics[width=1.\linewidth]{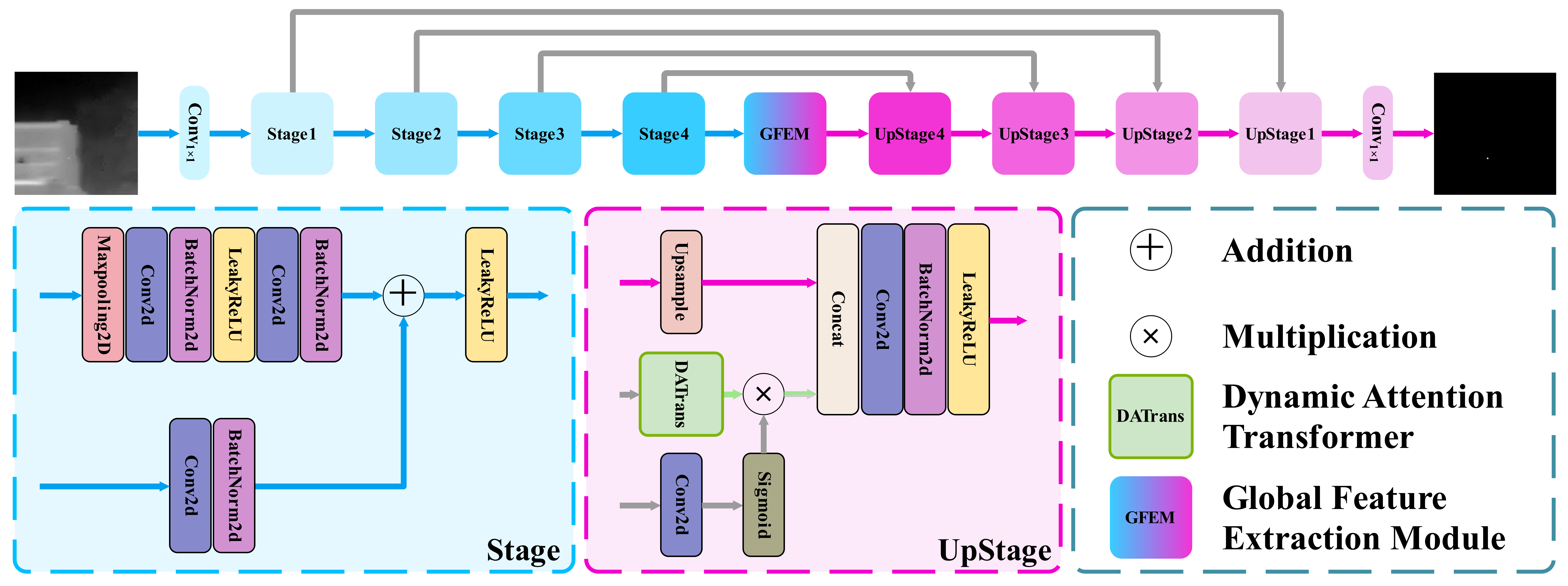}
    \caption{The overall structure of DATransNet. The red represents the upsampling stage (UpStage), and the blue corresponds to the downsampling stage (Stage).}
    \label{backbone}
\end{figure*}

\begin{enumerate}
    \item We propose DATransNet, which utilizes DATrans to extract detailed information by simulating the central difference convolution (CDC) with dynamic weights.
    \item We introduce GFEM to incorporate global information into our network.
    \item We conducted comparative experiments on the IRSTD-1K and NUDT-SIRST datasets, and the results show that DATransNet outperforms many existing methods.
\end{enumerate}

\begin{figure}
    \centering
    \includegraphics[width=.9\linewidth]{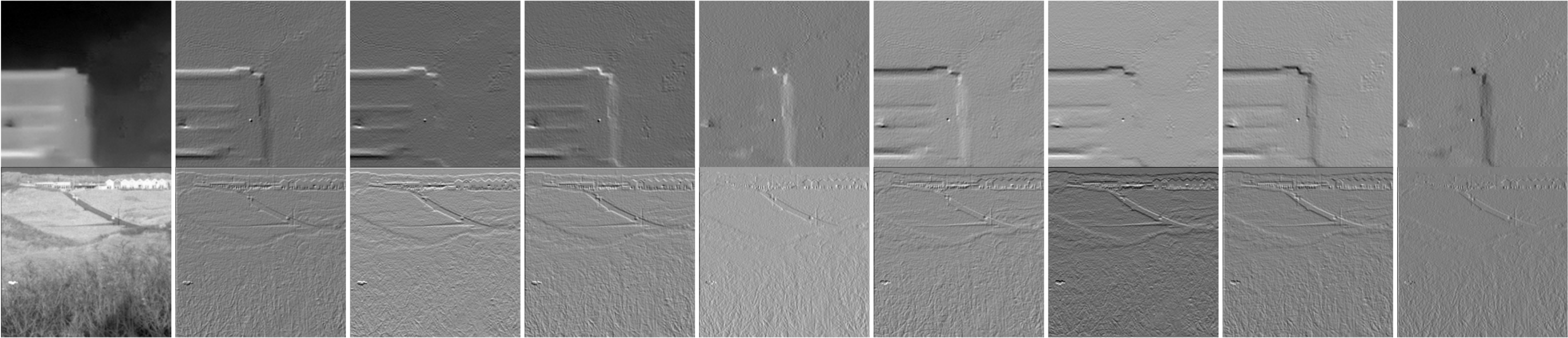}
    \caption{The first column is the original image, and the following 8 columns are the results of the original image using different edge convolutions across various directions.}
    \label{CDC_label}
\end{figure}

\section{Methods}
The overall architecture of DATransNet is illustrated in Fig. \ref{backbone}. 
DATransNet is based on U-Net and implements DATrans and GFEM to extract detailed information and contextual features, respectively.

\subsection{Dynamic Attention Transformer (DATrans)}
\label{sec:gtrans}
As shown in Fig. \ref{CDC_label}, the targets in ISTD are small and dim and do not have complex texture information. 
Therefore, the details in the image are vital for ISTD.
An effective approach to extracting the details is to take advantage of the differences between local pixels and their surroundings, such as CDCs. 
CDCs are defined in Eq. \ref{eq_0}.
\begin{equation}
out_{\hat{c}xy} = \sum^{c_{inp}-1}_{i=0} \sum^{7}_{j=0} w_{\hat{c}ijxy} (b_{ijxy} - o_{ixy}) \\
\label{eq_0}
\end{equation}
Here, \(c_{inp}\) represents the number of input channels, \(o_{ixy}\) denotes the input value at the \(i\)-th channel and position \((x,y)\), and \(b_{ijxy}\) refers to the \(j\)-th surrounding value around \(o_{ixy}\).
\(out_{\hat{c}xy}\) denotes the output of the \(\hat{c}\)-th channel at position \((x, y)\).  \(w_{\hat{c}ijxy}\) is the weight applied to the difference \(b_{ijxy} - o_{ixy}\).

The importance of different edge-convolution results varies from image to image. 
As shown in Fig. \ref{CDC_label}, the target in the upper image of the last column is clearly distinct, while that in the bottom row appears blurry.
So, the importance of difference results in various directions is varied for different images.
The weights $w_{\hat{c}ijxy}$ of the CDCs should be dynamically adjusted in response to changes in the input images.

We divide Eq. \ref{eq_0} into a fusion of Eq. \ref{eq_1_0}, Eq. \ref{eq_1}, and Eq. \ref{eq_2}. 

\begin{figure}
    \centering
    \includegraphics[width=1.\linewidth]{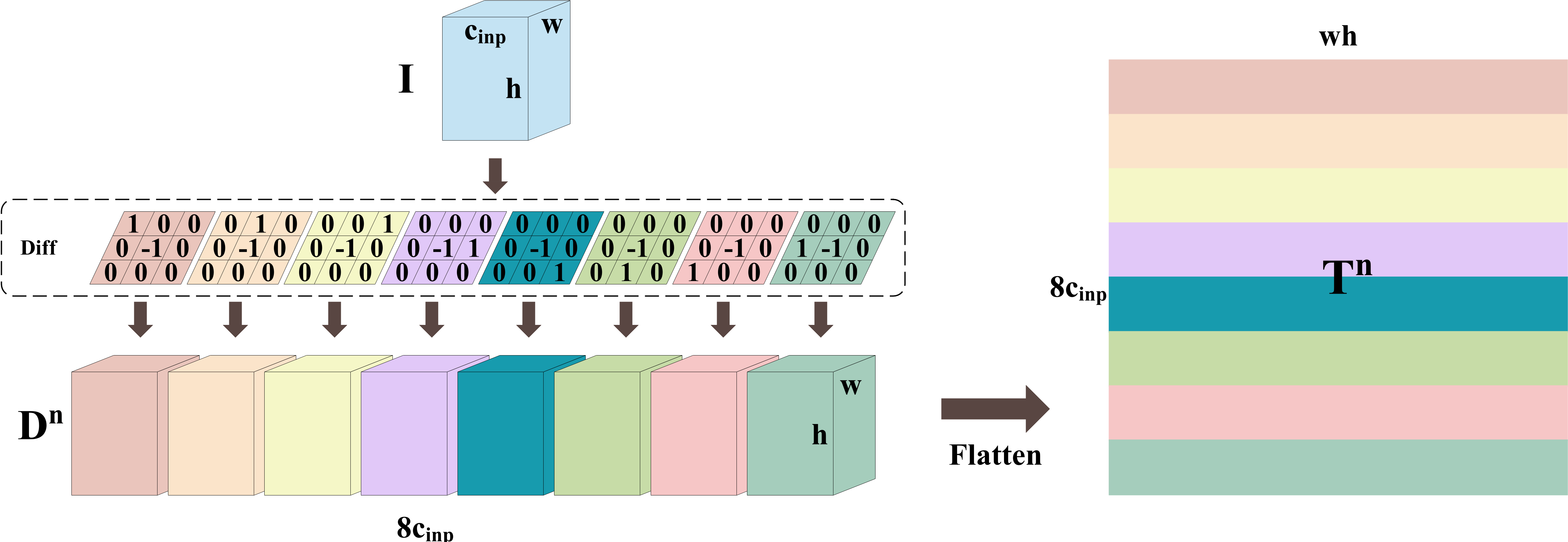}
    \caption{$T^n$ is derived by applying edge detection convolutions with a dilation ratio of $n$ to the image $I$, followed by flattening the output.}
    \label{fig_1}
\end{figure}

Eq. \ref{eq_1_0} and Eq. \ref{eq_1} describe the process of extracting difference features \( T^n \in \mathbb{R}^{8c_{inp} \times (wh)} \), as shown in Fig. \ref{fig_1}. We apply edge convolutions with a dilation ratio of $n$ to get \( D^n \in \mathbb{R}^{8c_{inp} \times w \times h} \) from the input image \( I \in \mathbb{R}^{c_{inp} \times w \times h} \), as outlined in Eq. \ref{eq_1_0}. The $Diff$ refers to performing eight edge convolutions on the input image. In Eq. \ref{eq_1}, the \( Flatten \) reshapes \( D^n \) into \( T^n \). In addition, \( w \) represents the input width, \( h \) denotes the input height, and \( c_{inp} \) denotes the number of input channels. In the process, we convert the spatial differences between eight neighborhoods to the dimension of the channel.      

Then we use the weight matrix \( M_w \in \mathbb{R}^{c_o \times 8c_{inp}} \) to obtain the output \( Out \in \mathbb{R}^{8c_{inp} \times (wh)} \), as shown in Eq. \ref{eq_2}, where \( Reshape \) refers to the transformation process transforming $M_wT^n$ into $Out^n \in R^{c_o\times w\times h}$ and \( c_o \) is the number of output channels. 

\begin{equation}
    D^n = Diff(I)
    \label{eq_1_0}
\end{equation}
\begin{equation}
    T^n = Flatten(D^n)
    \label{eq_1}
\end{equation}
\begin{equation}
    Out^n = Reshape(M_wT^n)
    \label{eq_2}
\end{equation}

\begin{figure}
    \centering
    \includegraphics[width=1.\linewidth]{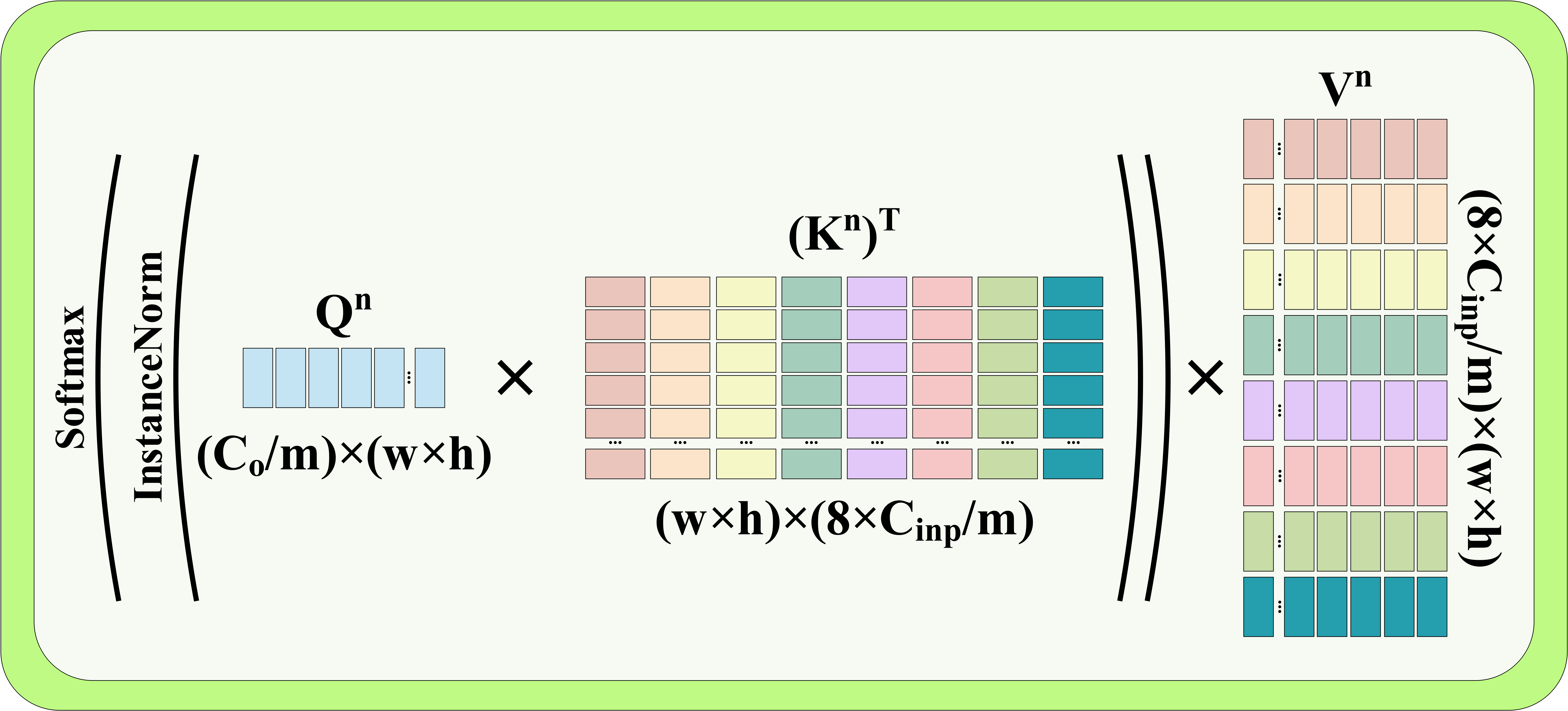}
    \caption{The overall structure of DATrans. }
    \label{fig_0}
\end{figure}

When $M_w$ in Eq. \ref{eq_2} changes with the input image $I$, the CDCs may also vary accordingly. Based on the analysis, we introduce the DATrans to simulate CDCs with dynamic weights for different directions. The structure of DATrans is shown in Fig. \ref{fig_0}.

To improve the detection capability for a variety of targets, we utilize varying dilation ratios across different heads in DATrans. For each head, the difference feature ($T^n$) extraction process follows the same approach as that described in Eq. \ref{eq_1_0} and Eq. \ref{eq_1}.
We treat each channel with 8 surrounding information as a token.

Furthermore, the input \( I \) is reshaped to form \( O \in \mathbb{R}^{c_{inp} \times wh} \), as shown in Eq. \ref{reshape}. 
\( O \) serves as the query token. The key and value tokens come from \( T^n \in \mathbb{R}^{8c_{inp} \times wh} \).
\begin{equation}
    O = Flatten(I)
\label{reshape}
\end{equation}
\begin{align}
    Q^n &= W^n_QO, K^n = W^n_KT^n, V = W^n_VT^n
\end{align}
where $W^n_Q\in \mathbb{R}^{\frac{c_o}{m} \times c_{inp}}$, $W^n_K \in \mathbb{R}^{\frac{8c_{inp}}{m} \times 8c_{inp}}$, $W^n_V \in \mathbb{R}^{\frac{8c_{inp}}{m}\times 8c_{inp}}$ are the learnable weights and $m$ denotes the number of heads. 
We use $Q^n\in \mathbb{R}^{\frac{c_o}{m}\times wh}$ and $K^n\in  \mathbb{R}^{\frac{8c_{inp}}{m}\times wh}$ to produce the attention matrix $M^n\in \mathbb{R}^{\frac{c_o}{m}\times \frac{8c_{inp}}{m}}$. 
$V^n\in \mathbb{R}^{\frac{8c_{inp}}{m}\times wh}$ is multiplied by $M^n$.
The result $Out^n \in \mathbb{R}^{\frac{c_o}{m}\times wh}$ is defined as follows:
\begin{equation}
    \begin{aligned}
    Out^n & = Softmax(Norm(\frac{Q^n(K^n)^T}{w})) V^n \\
    & = M^n V^n= (M^n W_{V}^{n}) D^{n} = M^n_{mix} D^{n}
    \label{end}
    \end{aligned}
\end{equation}
where \( M^n_{mix}\in\mathbb{R}^{\frac{c_o}{m}\times 8c_{inp}}  \) denotes the dynamic matrix derived from an attention matrix \( M^n \) and a learnable weight matrix \( W_V^n \). $Norm$ refers to Instance Normalization, which normalizes the similarity matrix for each instance on the similarity maps, ensuring smooth gradient propagation. According to Eq. \ref{eq_2} and \ref{end},  $Out^n$ is the equivalent result of the CDC with dynamic weights. 

At last, we concatenate the results of varied heads which have different dilation ratios, reshape them and use a $1\times 1$ convolution to fuse them to get the final output $Out$. In addition, $m$ is the number of heads.
\begin{equation}
    Out = Conv_{1\times 1}(Reshape(Out^1,...,Out^m))
\end{equation}

\subsection{Global Feature Extraction Module (GFEM)}
\begin{figure} 
    \centering
    \includegraphics[width=1.\linewidth]{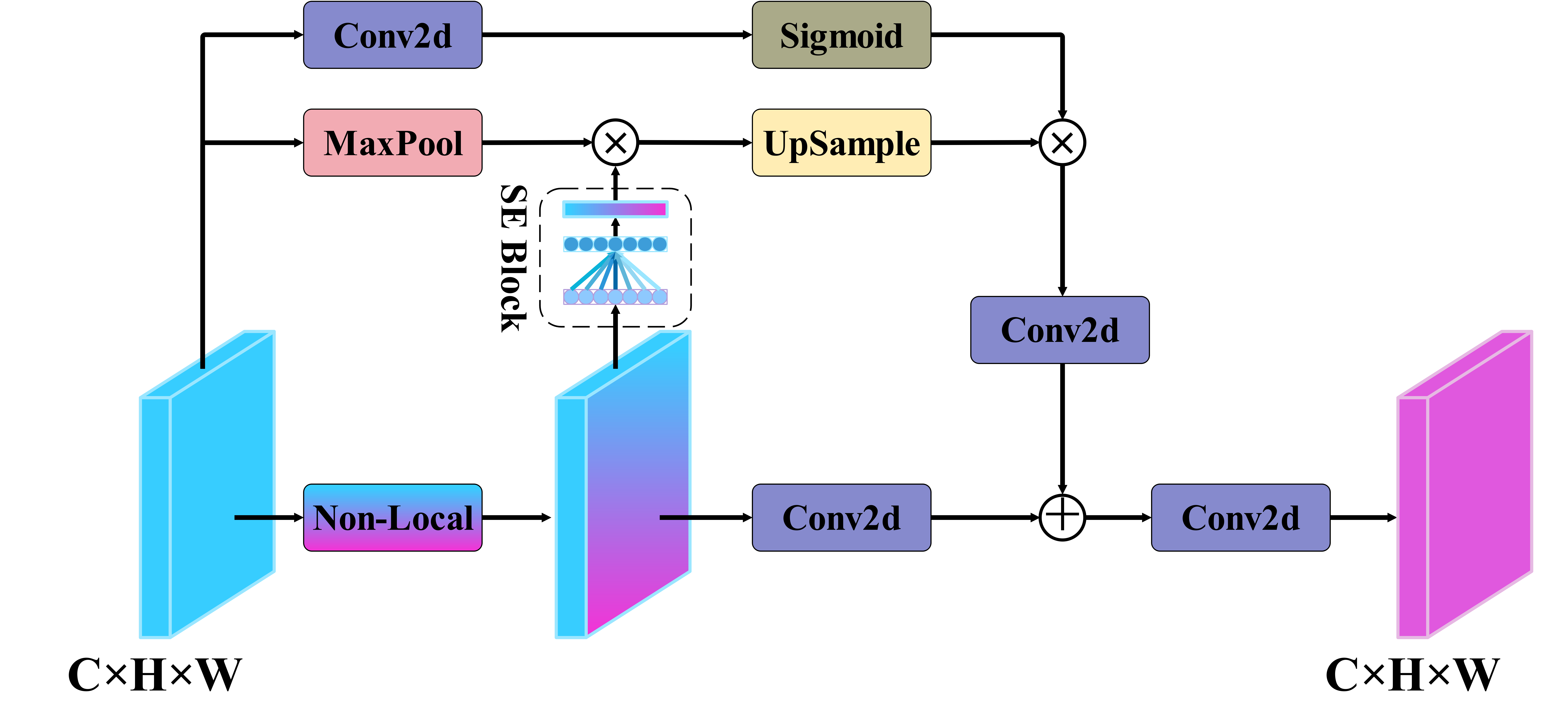}
    \caption{The structure of the GFEM using the Non-local (Non-local Attention Module) to capture global spatial features and SE Block (Squeeze-and-Excitation Block) to extract global channel information.}
    \label{fig_3}
\end{figure}

Background information is crucial alongside the detailed features of small targets in ISTD. 
However, background information is based on a global perspective of the whole image.
So, we propose the GFEM, as illustrated in Fig. \ref{fig_3}.
In GFEM, we incorporate the attention mechanism to provide our model with a broader perspective. This module includes three key steps: First, we utilize the non-local attention mechanism \cite{non_local} to compute spatial attention over the deepest feature map, which is significantly smaller in size compared to the input image. 
Initially, we apply a \( 1 \times 1 \) convolution to reduce the number of channels from \( c \) in \( X \in \mathbb{R}^{ c \times w\times h} \) to \( c' \) in \( Q \in \mathbb{R}^{ c' \times w\times h}\), \( K \in \mathbb{R}^{ c' \times w\times h} \), and \( V \in \mathbb{R}^{ c' \times w\times h} \). 
Besides, $w$ and $h$ denote the width and height of the feature map input to the non-local module. 
Then, we reshape the feature maps and compute \( Y \in \mathbb{R}^{ c \times w \times h} \) according to the following equation:
\[
Y = Conv_{1 \times 1}(Reshape(V' \cdot Softmax(Q' \cdot (K')^T)))
\]
where \( Q' \in \mathbb{R}^{ c' \times (wh)} \), \( K' \in \mathbb{R}^{ c' \times (wh)} \), and \( V' \in \mathbb{R}^{ c' \times (wh)} \) are the results of reshaping \( Q \), \( K \), and \( V \). \( Softmax \) is softmax layer, and \( Conv_{1 \times 1} \) is a convolution with a \( 1 \times 1 \) kernel, to increase the number of channels. 
Second, we incorporate a squeeze and excitation block \cite{senet} to enhance channel-wise perception. Finally, we concatenate the results from the global spatial and channel attention modules and employ convolution to fuse them, enabling GFEM to capture comprehensive feature representations.

Subsequently, the output feature map acquires both spatial and channel global receptive fields, which are vital for effective object detection. 

\subsection{Loss Function}
The loss function utilized in our model training is the soft intersection over union (SoftIoU) loss, as shown in Eq. \ref{softiou}.
\begin{equation} 
loss = 1 - \frac{\sum_{i,j} p_{i,j} \cdot g_{i,j}}{\sum_{i,j} p_{i,j} + \sum_{i,j} g_{i,j} - \sum_{i,j} p_{i,j} \cdot g_{i,j}} 
\label{softiou} 
\end{equation}
where $g_{i,j}$ and $p_{i,j}$ denote the ground truth and the output of our network at the $(i,j)$, respectively.

\begin{figure*}
    \centering
    \includegraphics[width=1.\linewidth]{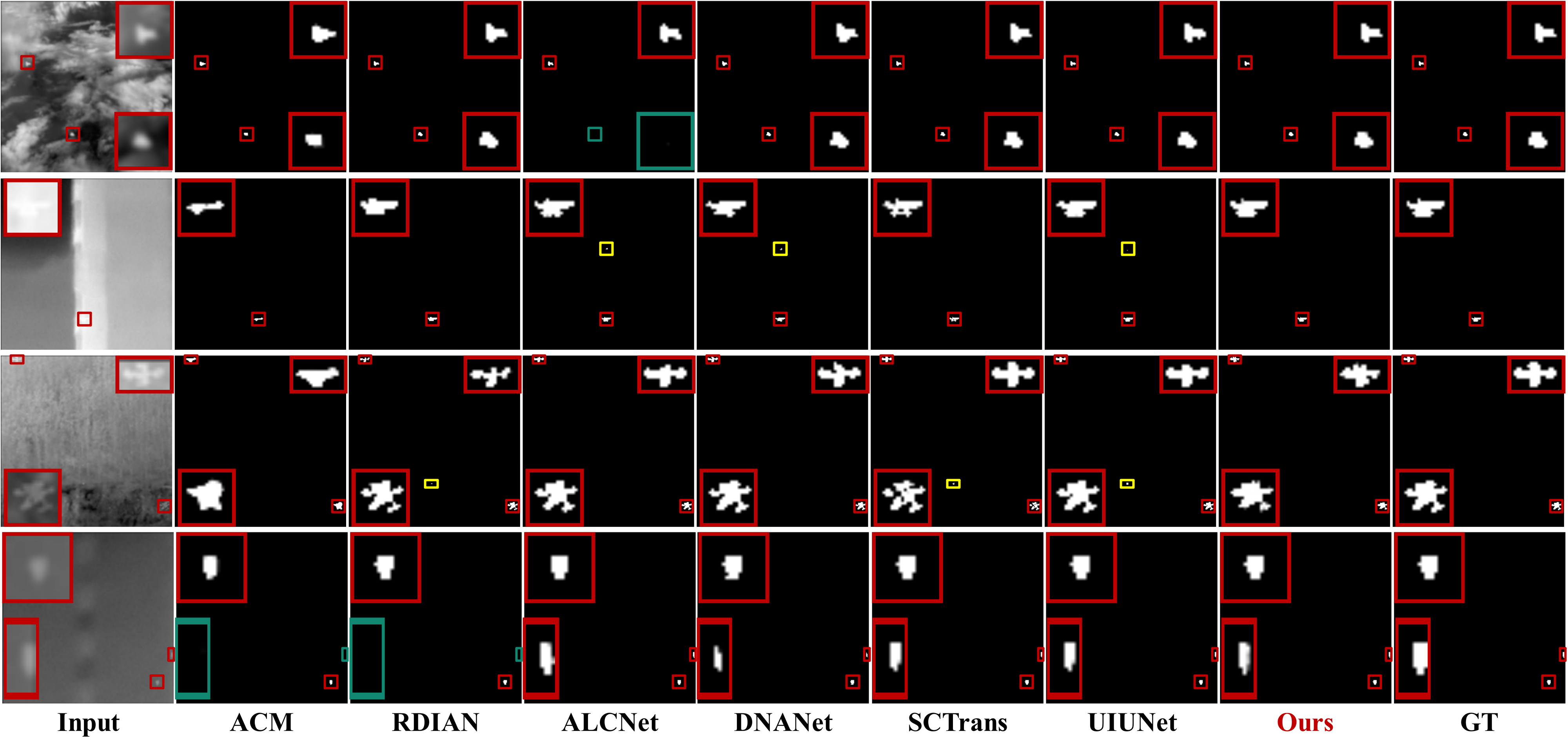}
    \caption{Visual results from varied data-driven methods. The red, green and yellow boxes represent detected targets, missed targets, and false alarms.}
    \label{comparsion_fig}
\end{figure*}

\section{Experiment and Analysis}

This section describes the experimental details and the evaluation metrics, followed by a series of ablation studies to verify the proposed modules. Finally, a comprehensive comparative experiment with other methods on qualitative and quantitative results demonstrates that our approach outperforms existing state-of-the-art (SOTA) methods.

\subsection{Dataset and Evaluation Metrics}
The datasets used for evaluating the DATransNet are NUDT-SIRST\cite{li2022dense} and IRSTD-1K\cite{zhang2022isnet}. 
The NUDT-SIRST dataset consists of 1327 images whose resolution is 256×256, with a split of 332 images for testing, 332 for validation, and 663 for training.
The IRSTD-1K dataset contains 1001 images with a resolution of 512×512. We select 101 images for testing, 100 images for validation, and 800 images for training.
The evaluation metrics include mean intersection over union ($mIoU$), F1-measure ($F_1$), the probability of detection ($P_d$), and false alarm rate ($F_a$). 

\subsection{Experimental Environment and Parameter Settings}

All models are built on the PyTorch framework and are trained on an NVIDIA GeForce RTX 4080 GPU. We employ the Adam optimizer, starting with an initial learning rate of \(5 \times 10^{-4}\). This learning rate is reduced to \(5 \times 10^{-5}\) at epoch 200 and further decreases to \(5 \times 10^{-6}\). The batch size is 4, and there are 400 epochs.


\subsection{Ablation Study}

\subsubsection{\textbf{Studies of Module-wise Performance Gain}} 
In this ablation study, we begin with the baseline. Then, we test the performance of the DATransNet's module, and the results are shown in Tab. \ref{component}. The integration of DATrans and GFEM leads to a progressive enhancement in the performance of our network..

\begin{table}
  \setlength{\abovecaptionskip}{0cm}  
  \renewcommand\arraystretch{1}
  \footnotesize
  \centering
  \vspace{-1\baselineskip}
  \caption{\centering Studies Of Different Components on NUDT-SIRST (The best results are \textbf{bold})}
  \label{tab:IRSTD-ablation-module}
  \setlength{\tabcolsep}{2pt}
  \begin{tabular}{c|ccccc}
  \hline
      \multicolumn{1}{c|}{Module}  
      &\small{mIoU(\%)} & \small{$F_1(\%)$} & \small{$P_d(\%)$} &\small{$F_a(10^{-6})$}\\
  \hline
                 U-Net & 91.31&95.44 &97.98 &4.46  \\
                 U-Net+DATrans &\underline{94.25} &\underline{97.03} &\underline{98.83} &\underline{2.73}\\
                 U-Net+GFEM &92.32 &96.14 &96.30 &3.96 \\
                 U-Net+DATrans+GFEM &\textbf{94.93} &\textbf{97.39} &\textbf{99.04} &\textbf{2.00} \\
  \hline
\end{tabular}
\label{component}
\end{table}

\subsubsection{\textbf{Studies of Dilation Rate for DATrans}} 
As mentioned in Section \ref{sec:gtrans}, DATrans employs varying dilation rates across different detection heads. Tab. \ref{Dilation} demonstrates the efficacy of this strategy, as models with diverse dilation ratios outperform those with a single rate. Furthermore, networks with dilation ratios of 1 and 3 yield superior results.

\begin{table}[htbp]
  \setlength{\abovecaptionskip}{0cm}  
  \renewcommand\arraystretch{1}
  \footnotesize
  \centering
  \caption{\centering Studies on different dilation rates on NUDT-SIRST (The best results are \textbf{bold})}
  \label{tab:IRSTD-ablation-Dilation}
  \setlength{\tabcolsep}{2pt}
  \begin{tabular}{c|ccccc}
  \hline
      \multicolumn{1}{c|}{Dilation Rate}   
      &\small{mIoU(\%)} & \small{$F_1(\%)$} & \small{$P_d(\%)$} &\small{$F_a(10^{-6})$}\\
  \hline
                 1  &93.53 &96.30 &\underline{98.89} &5.58  \\
                 1,2 &94.41 &97.12 &98.83 &2.46\\
                 1,3 &\textbf{94.93} &\textbf{97.39} &\textbf{99.04} &\underline{2.00} \\
                 1,5 &94.24 &97.03 &98.65 &\textbf{1.47} \\
                 1,2,3,4&\underline{94.58} &\underline{97.20} &98.04 &2.21\\
  \hline
\end{tabular}
\label{Dilation}
\end{table}

\begin{table}[htbp]
  \setlength{\abovecaptionskip}{0cm}  
  \renewcommand\arraystretch{1}
  \footnotesize
  \centering
  \vspace{-1\baselineskip}
  \caption{\centering Studies on GFEM on NUDT-SIRST (The best results are \textbf{bold})}
  \setlength{\tabcolsep}{2pt}
  \begin{tabular}{cc|cccccc}
  \hline
      \small{Non-local}
      &\small{SE Block}
      &\small{Params(M)}
      &\small{GFLOPs(G)}
      &\small{mIoU(\%)} 
      &\small{$F_1(\%)$} \\
  \hline
                $-$ & $-$ & \textbf{3.70}&\textbf{10.82} &94.25 &97.03\\
                $\checkmark$ & $-$ &4.03 &\underline{10.89} &\underline{94.69} &\underline{97.27}  \\
                 $-$ &$\checkmark$ &\underline{4.02} &\underline{10.89} &94.53 &97.19 \\
                 $\checkmark$ &$\checkmark$ &4.04 &10.90 &\textbf{94.93} &\textbf{97.39}  \\
  \hline
\end{tabular}
\label{GFEM}
\end{table}
\begin{table*}
  \setlength{\abovecaptionskip}{0 cm}  
  \renewcommand\arraystretch{1}
  \footnotesize
  \centering
  \vspace{-1\baselineskip}
  \caption{Quantitative Comparsion With Different Methods on NUDT-SIRST and IRSTD-1K (The best results are \textbf{bold}, second best results are underline.)}
  \label{tab:Compare-IRSTD-ablation}
  \setlength{\tabcolsep}{8pt}
  \begin{tabular}{c|c|cccc|cccc}
  \hline
      \multicolumn{1}{c|}{Model} & \multicolumn{1}{c|}{Metrics}  & \multicolumn{4}{c|}{NUDT-SIRST}  & \multicolumn{4}{c}{IRSTD-1K}    \\
      &\small{Params(M)}
      &\small{mIoU(\%)} & \small{$F_1(\%)$} & \small{$P_d(\%)$} &\small{$F_a(10^{-6})$}
      &\small{mIoU(\%)} & \small{$F_1(\%)$} & \small{$P_d(\%)$} &\small{$F_a(10^{-6})$} \\
  \hline
  \vspace{.5\baselineskip}
                 ACM\cite{dai2021asymmetric}     &\underline{0.40}&70.97&82.99&97.67&7.33 &64.09&78.11&88.55&17.21\\
                 RDIAN\cite{sun2023receptive}   &0.90&87.78&93.47&97.88&9.67 &65.68&79.30&\underline{91.24}&\underline{10.31}\\
                 ALCNet\cite{dai2021attentional}  &\textbf{0.37}&92.45&96.08&98.94&2.62 &65.05&78.60&90.57&19.42\\
                 DNANet\cite{li2022dense}  &4.69&93.73&96.70&\textbf{99.25}&4.55 &66.73&80.08&89.22&\textbf{7.97}\\
                 UIU-Net\cite{wu2022uiu} &50.54&\underline{94.11}&\underline{96.96}&97.98&\textbf{0.74} &\underline{67.94}&\underline{80.90}&90.57&24.44\\
                SCTransNet\cite{Yuan2024SCTransNetSC} &11.19 &93.83 &96.77 &98.20 &\underline{1.56} &65.83 &79.40 &90.57 &14.50\\
                
                 Ours    &4.04&\textbf{94.93} &\textbf{97.39} &\underline{99.04} &2.00 &\textbf{68.56} &\textbf{81.34} &\textbf{93.60} &24.96\\
  \hline
  
\end{tabular}
\label{Comparsion}
\vspace{-2\baselineskip}
\end{table*}
\subsubsection{\textbf{Studies of GFEM}}
As mentioned in \ref{GFEM}, we conduct the experiments on the modules in the GFEM. The network combining the two modules performs better with a slight increase in computational complexity.

\subsection{Comparsion with State-of-the-art (SOTA) Methods}
We compare it with several SOTA methods on NUDT-SIRST and IRSTD-1K, including SCTransNet \cite{Yuan2024SCTransNetSC}, UIUNet \cite{wu2022uiu}, ACM \cite{dai2021asymmetric}, ALCNet \cite{dai2021attentional}, AGPCNet \cite{zhang2023attention}, RDIAN \cite{sun2023receptive}, and DNANet \cite{li2022dense}. In addition, deep supervision is not employed to ensure equality in any of the networks. We used DATransNet with dilation ratios of 1 and 3 for comparative evaluation. The results of the experiments include qualitative and quantitative results.

\subsubsection{\textbf{Qualitative Results}}
As shown in Fig. \ref{comparsion_fig}, our model preserves the shape of targets more closely to ground truth with low missed detections and false alarms.


\subsubsection{\textbf{Quantitative Results}}

As shown in the Tab. \ref{Comparsion}, our method achieves impressive performance, including a $mIoU$ of $94.93\%$, an $F1$ score of $97.39\%$, and a $Pd$ of $99.04\%$, a $F_a$ of $2.00 \times 10^{-6}$. Similarly, on the IRSTD-1K dataset, our method is a leading network. 

\begin{figure}[htbp]
  \centering
  \subfloat[]{
    \includegraphics[width=.2\textwidth]{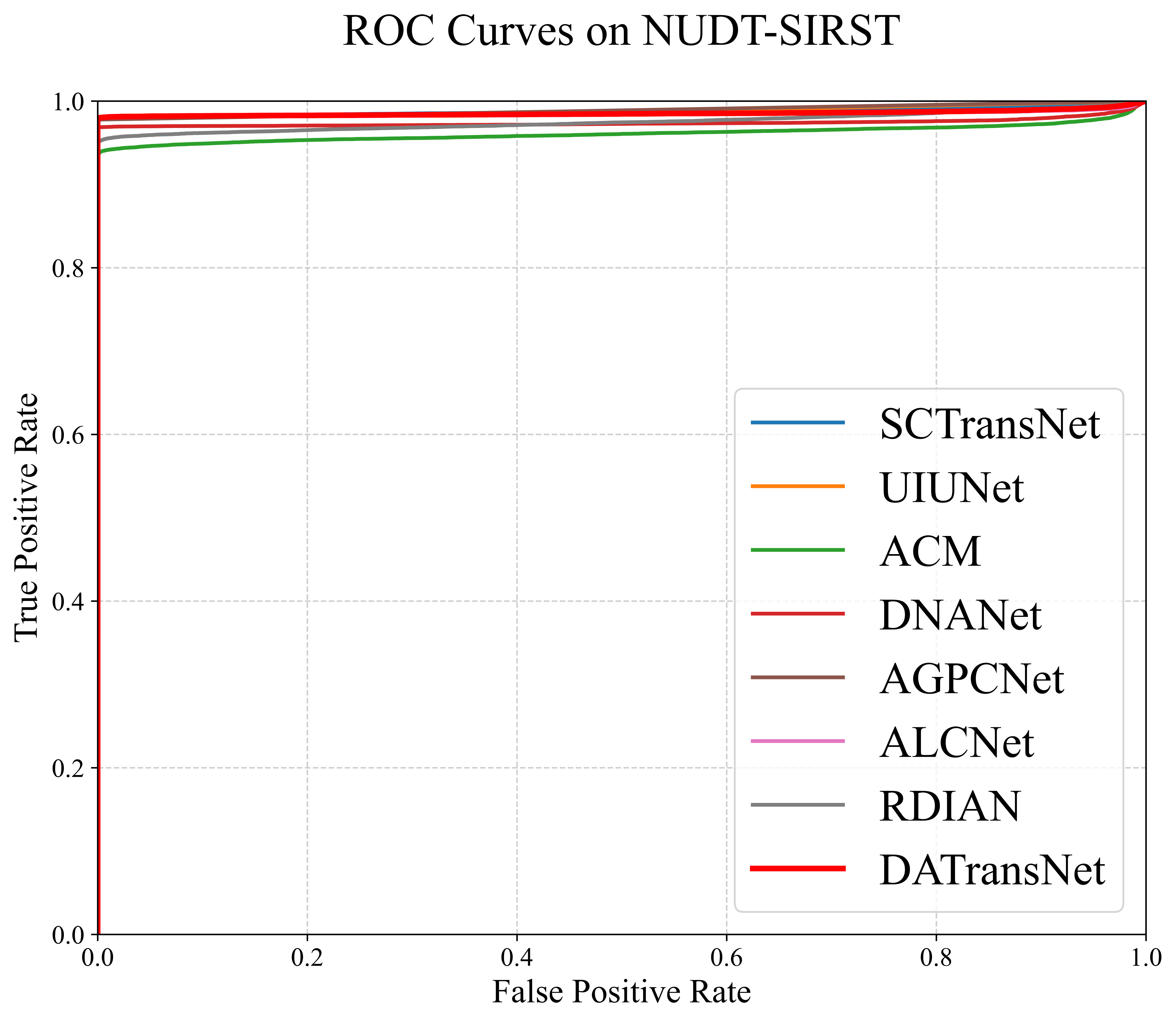}
  }
  \subfloat[]{
    \includegraphics[width=.2\textwidth]{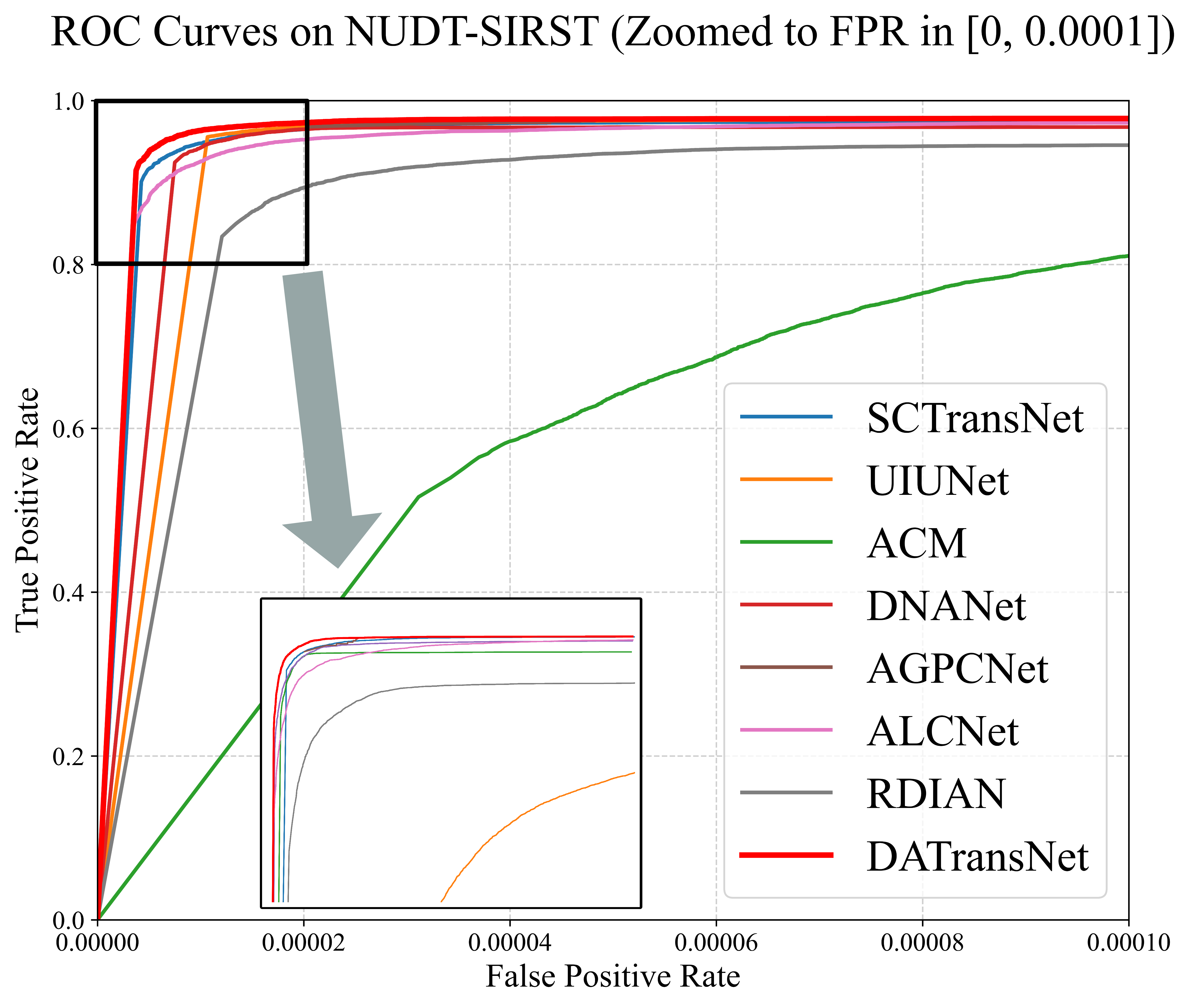}
  }
  \\
  \subfloat[]{
    \includegraphics[width=.2\textwidth]{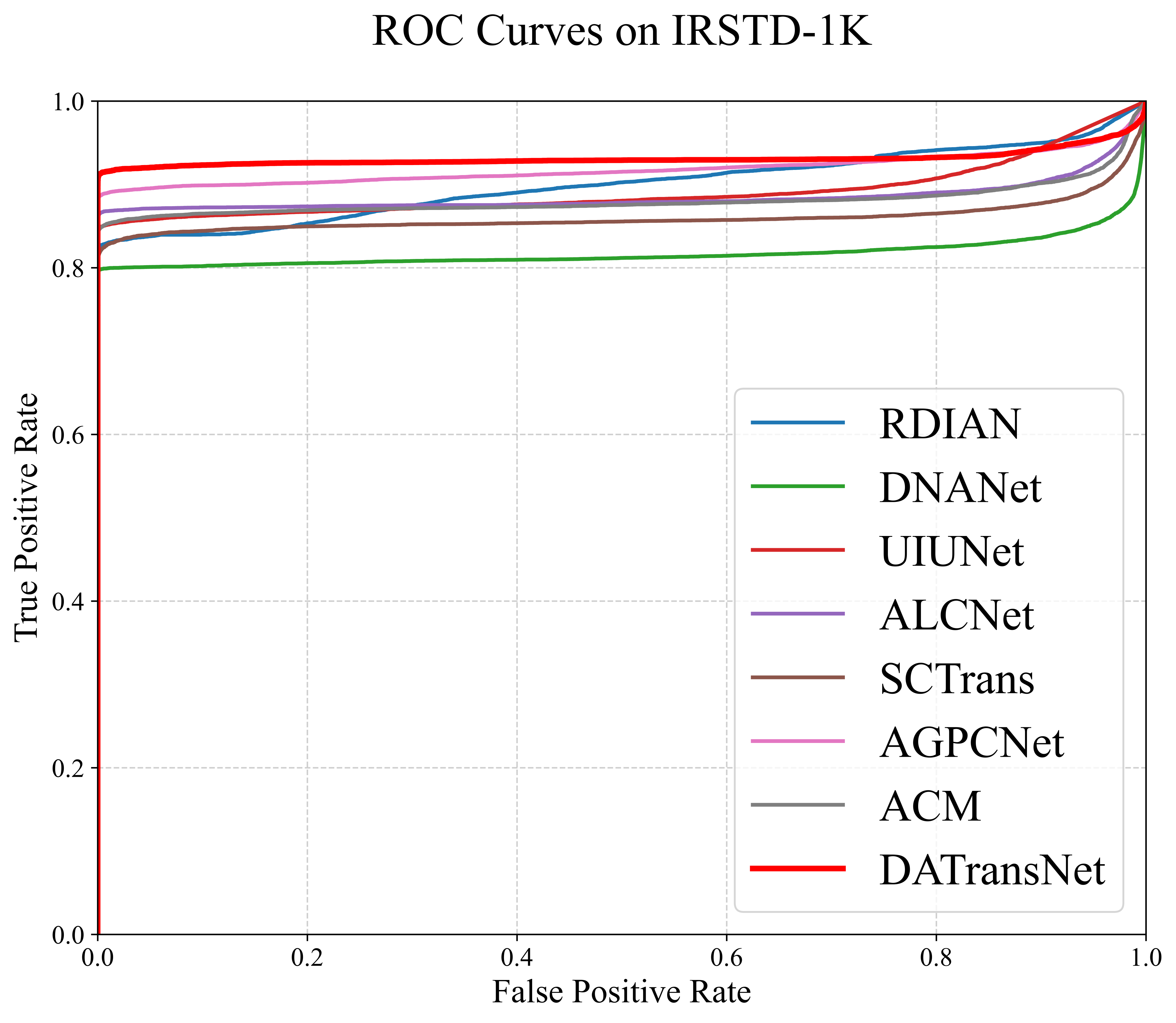}
  }
  \subfloat[]{
    \includegraphics[width=.2\textwidth]{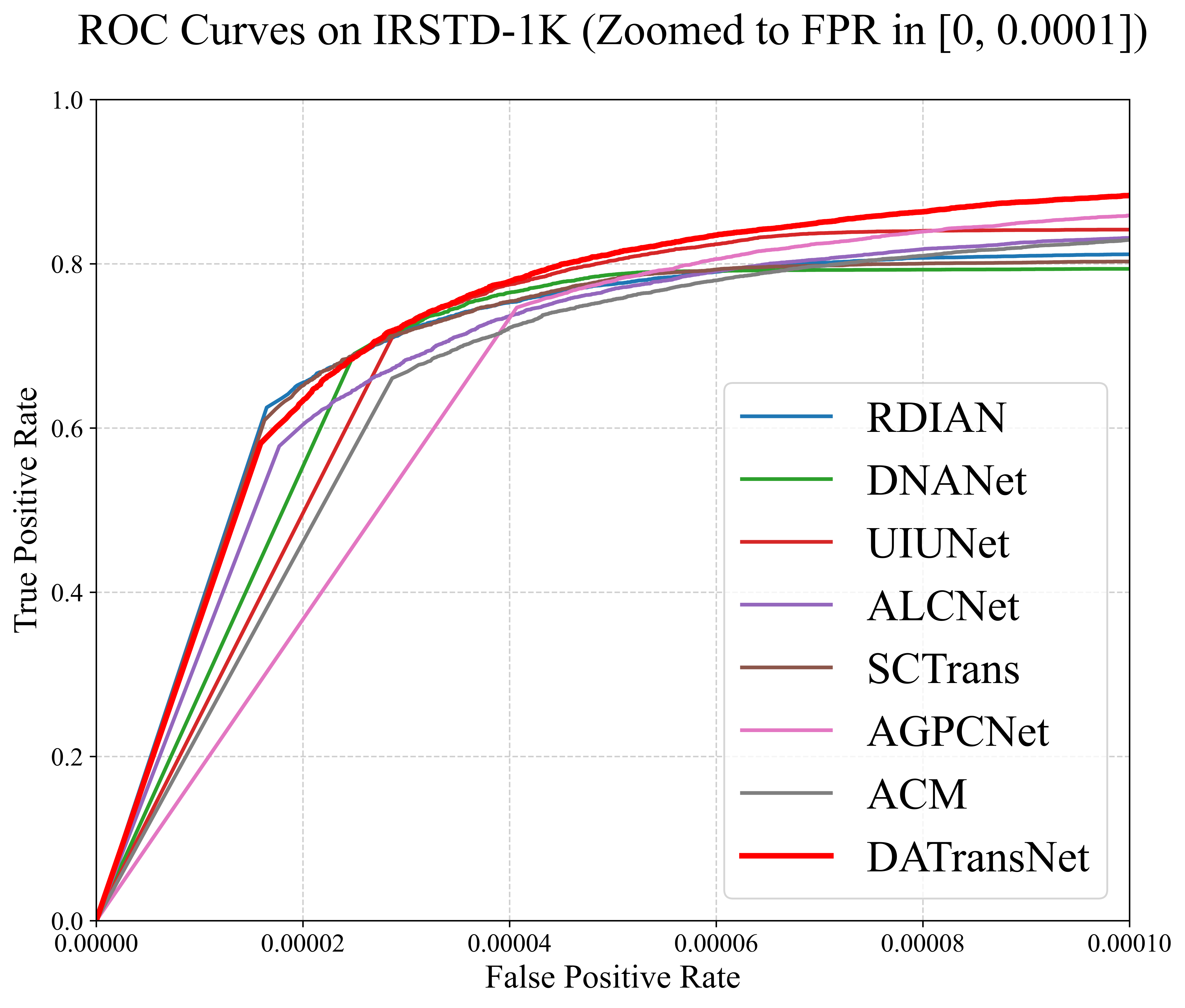}
  }
  \caption{ROC curves of varied networks on NUDT-SIRST and IRSTD-1K datasets.}
  \label{ROC}
\end{figure}
Furthermore, we used the ROC curve to analyze the performance of various models, as depicted in Fig. \ref{ROC}.  The model proposed in this article consistently shows good performance in both ROC and $mIoU$.


\section{Conclusion}


In this Letter, we propose an ISTD network that enhances detection performance.  DATrans was proposed to enhance the network's local gradient feature extraction and detection performance. Additionally, we propose the GFEM, which focuses on global perception across the entire feature map and learns relationships between distant pixels. 
We have conducted extensive experiments, indicating satisfied results with fewer parameters. 
The model is still a black-box approach because its capability for dynamic weight adjustments lacks mathematical proof. Moreover, it suffers from false alarms when processing high-noise images. 
To address these limitations, we aim to improve the model in two ways. At first, we plan to improve the interpretability of the network. Then, we would like to integrate temporal information into the model to improve its robustness.

\bibliographystyle{IEEEtran}     
\bibliography{ref}                     
\end{document}